# Comparison of Transfer Learning based Additive Manufacturing Models via A Case Study

**Yifan Tang**
Email: yta88@sfu.ca

**M. Rahmani Dehaghani**
Email: mra91@sfu.ca

**G. Gary Wang**[1]
Email: gary_wang@sfu.ca

Product Design and Optimization Laboratory, Simon Fraser University, Surrey, BC, Canada

**Abstract**

Transfer learning (TL) based additive manufacturing (AM) modeling is an emerging field to reuse the data from historical products and mitigate the data insufficiency in modeling new products. Although some trials have been conducted recently, the inherent challenges of applying TL in AM modeling are seldom discussed, e.g., *which source domain to use, how much target data is needed,* and *whether to apply data preprocessing techniques.* This paper aims to answer those questions through a case study defined based on an open-source dataset about metal AM products. In the case study, five TL methods are integrated with decision tree regression (DTR) and artificial neural network (ANN) to construct six TL-based models, whose performances are then compared with the baseline DTR and ANN in a proposed validation framework. The comparisons are used to quantify the performance of applied TL methods and are discussed from the perspective of similarity, training data size, and data preprocessing. Finally, the source AM domain with larger qualitative similarity and a certain range of target-to-source training data size ratio are recommended. Besides, the data preprocessing should be performed carefully to balance the modeling performance and the performance improvement due to TL.

## 1 Introduction

Additive manufacturing (AM) modeling aims to reveal the process-structure-property correlations, which are then applied widely in AM process optimization and control to improve the final product qualities [1]. Recently, data-driven modeling methods have attracted increasing attention in AM due to their ability to reach a balance among model transferability, prior knowledge requirement, and experiment requirement [2–4]. These methods extract information from experiments and/or simulations to estimate AM process behaviors by various machine learning (ML) techniques, such as support vector regression (SVR) for product geometry prediction [5], Gaussian process regression (GPR) for melt pool geometry estimation [6], recurrent neural network (RNN) for thermal field modeling [7], convolutional neural network (CNN) for bead geometry control [8], etc. Although these ML-based models have potential applications in *in-situ* monitoring/control and process optimization, their online prediction accuracy is restricted by the data size in offline training. In other words, when the offline training data is insufficient, the constructed ML models could not provide an acceptable prediction for online conditions. To reduce the impact of the data insufficiency problem and further improve the modeling performance, sharing data among different infrastructures or processes is highlighted in references [2,3]. The idea of sharing data is the same as transfer learning (TL), which aims to improve the learning performance of the target task with the assistance of the knowledge in the source task [9].

Recently, applications of different TL methods are increasing in AM modeling gradually, due to their ability to bridge various data sources and no requirement of large data sets. These applications could be classified according to different types of TL methods, whose brief introductions are presented below. More details about state-of-the-art research on TL-based AM modeling could refer to our recent literature review paper [10].

- Instance-base TL

The instance-based TL applications select partial source data and reuse them directly in constructing target models, which means the source and target AM modeling problems have identical input and output variables. For instance, the Two-stage TrAdaBoost.R2 integrated GPR model is proposed to predict line geometries (e.g., width, thickness, and edge roughness) of aerosol jet printing (AJP) under various operating conditions [11]. Similarly, for the laser-based AM process optimization, the Bayesian updating framework is adopted to construct the target model with the increasing target data and the source data collected from published papers involving different steel powders and machines [12].

---
[1] Corresponding author

- Feature-based TL

The feature-based TL methods are a kind of data processing method, where the source and target features (defined based on input and output variables) are transformed into a common feature space to minimize their difference [13,14]. For example, a simplified transformation matrix integrated subspace alignment method is applied to tackle the data size disparity between source and target AJP modeling cases [11]. Then, the transformed source and target data are fed to the selected modeling framework to construct a target GPR model for geometry prediction.

- Model-based TL

Different from the above applications of instance/feature-based TL, more applications of the model-based TL are observed, where the target model construction and training process are facilitated with the source model structures and parameters. One idea is to construct the target model as a transformation of the pre-trained source model, where a function is learned to transform the target inputs into the source input space [15]. This framework has been applied widely to transfer in-plane or out-of-pane deviation knowledge among various AM products, such as laser powder bed fusion (LPBF) products fabricated with different materials (e.g., Ti-6Al-4V alloy and 316L stainless steel) [16], or with different printers (e.g., EOS M290 and MSU Renishaw AM 400) [17]. One similar work is proposed to predict geometries of printed lines based on AJP process parameters [11], where one global transformation vector is used to scale and shift the source response surface, and one local transformation vector is applied to map the target inputs to the source inputs. Both vectors are optimized simultaneously by minimizing the target prediction error with the genetic algorithm (GA).

The fine-tuning framework based on neural networks is another idea of model-based TL, which has been widely applied. By sharing partial pre-trained source model structure and parameters, this framework enables constructing target models with tailored input/output layers and training them with limited target data to obtain acceptable prediction performance. One group of its applications is the process and product property modeling, which explores the knowledge transferability within AM processes (e.g., from bead-on-plate process to bead-on-powder process [18], from LPBF to binder jetting process [19]), AM machines (e.g., from a Prima Power Laserdyne 430 3-axis CNC machine to a 5-axis DMD machine prototype [20]), AM products (e.g., stereolithography products with different geometries [21,22], fused decomposition modeling products with different metrology measurements [23]), and AM materials (e.g., from stainless steel 316L to bronze [24], from low-carbon steel to stainless steel 316L [25]). Different from the above works with AM-related source domains, the general-purpose and high-performance CNN backbones (e.g., AlexNet [26], ResNet [27], and VGG Net [28], GoogLeNet [29], EfficientNet [30], etc.) trained on the ImageNet [31] or COCO [32] database, have been tailored for various AM classification problems. Based on the limited sensor and digital AM images, some highly accurate classification models are developed and applied widely in defect detection, quality control, and *in-situ* monitoring of different AM processes [33–37].

- Multi-task learning

As an inductive TL method, multi-task learning (MTL) [38] relaxes the requirements on the size of source data, which differentiates itself from the above three TL methods. By using data in all/some tasks, MTL enables tackling several tasks simultaneously to improve the learning performance of each task. Based on this merit, this framework has been applied in geometry deviation prediction for various AM products (each of which has limited data) [39,40], and defect detection for products fimbriated with different printers and limited resources in a cloud platform [41]. However, its performance depends on the number of tasks (e.g., AM products) involved. If few tasks are available during the training, the learning performance would be poor as the data scarcity problem persists.

Although various TL methods have been applied with different pros and cons, they seldom discuss how to mitigate the inherent challenges of applying TL, e.g., *which source domain to use, how much target data is needed,* and *whether to apply data preprocessing techniques for a given AM modeling problem*. Consequently, when solving new problems, researchers would spend significant time finding the best choices via trial and error.

- *Which source domain to use?* The similarity between the source and target domains affects the TL performance greatly. If the similarity is low, the negative transfer would occur and damage the target learning [44]. Therefore, selecting an appropriate source domain is the prerequisite for applying TL.
- *How much target data is needed?* After selecting source/target domains and ML models, the TL performance depends on the source and target data sizes to some extent [45]. Especially in AM problems involving expensive experiments or simulations, the source data size is fixed based on completed source tasks, but the size of the target data to be generated is still uncertain for different problems.
- *Whether to apply data preprocessing techniques?* Considering that the data in real-world problems could have different magnitudes and types (e.g., continuous, discrete, or mixed variables), the data preprocessing techniques should be selected carefully according to the characteristics of ML methods; otherwise, the selected data preprocessing would deteriorate learning performances [46].

To explore the above three challenges, this paper designs a case study to discuss the effects of similarity, data size, and data preprocessing on TL performances in AM modeling problems. The remainder of this paper is structured as follows. The case study definition is presented with notations and data collection in Section 2. In Section 3, the structures of six reproduced TL-based models are discussed. Then the validation framework is designed in Section 4, based on which the performances of applied TL methods are discussed in Section 5. Finally, Section 6 provides a summary of this paper.

## 2 Case Study Definition

To clarify the designed case study, some related definitions are provided before discussions.

- *Domain $\mathcal{D}$*: The domain of any AM product is related to its manufacturing setting, such as printer machine, material, AM process, process parameters, etc. The "*input variable $x$*" in one domain is a set of parameters that affect product properties. A tuple of $n$ input variables is defined as $X = [x_1; ...; x_n]$.
- *Task $\mathcal{T}$*: Within one AM task, a product is to be fabricated with certain requirements, e.g., mechanical properties or *in-situ* manufacturing qualities. The "*output variable $y$*" refers to the property/quality of interest, whose relation with $x$ is formulated as $y = f(x|\theta)$, where $\theta$ contains all model parameters to be learned from a group of $(x, y)$ points. A tuple of $n$ outputs is presented as $Y = [y_1; ...; y_n]$.
- *Source/Target*: In AM, the source refers to available completed products, while the target refers to new products to be fabricated with the assistance of AM source knowledge.

Based on data from source AM products and fewer data from target AM products, the purpose of the case study is to *compare improvements in modeling performances caused by different TL methods* over the baseline target models. In this paper, the case study is designed based on the open-source dataset in [47], containing some process information (e.g., machine type, powder composition, laser power, laser speed, powder size, hatch spacing, layer thickness, etc.) and several target properties (e.g., relative density, microhardness, elastic-modulus, yield strength, etc.) about LPBF products fabricated with Ti-6Al-4V.

In the case study, one target task and two source tasks are defined according to different printers, as shown in Table 1. Intuitively, compared with source task 1 (EOS M270), source task 2 (SLM 250 HL) has a higher similarity with the target task (SLM 125 HL), as their machines are within the same series. In this paper, the relative density of the final LPBF product is selected as the interested output $y$. The laser power, laser speed, hatch spacing, and energy are selected as the input variables $x$, as they are the most influential factors [47]. Considering the different physical characteristics of machines, source and target tasks have different variation ranges for each process parameter, reflecting differences among different tasks. Finally, datasets $\boldsymbol{D}^{s1} = \{(\boldsymbol{x}_i^{s1}, y_i^{s1}) | i \in [1,49]\}$, $\boldsymbol{D}^{s2} = \{(\boldsymbol{x}_i^{s2}, y_i^{s2}) | i \in [1,32]\}$, and $\boldsymbol{D}^t = \{(\boldsymbol{x}_i^t, y_i^t) | i \in [1,24]\}$ are obtained for source task 1, source task 2, and target task respectively. In this paper, source data are only used for constructing the source model, while the target data is split to train and validate the target model.

Table 1 Definitions of source and target tasks in the case study

| Information | | Source task 1 | Source task 2 | Target task |
|---|---|---|---|---|
| Printers | | EOS M270 | SLM 250 HL | SLM 125 HL |
| Data size | | 49 | 32 | 24 |
| Range of process parameters | Laser power ($w$) | [40, 195] | [100, 375] | [50, 100] |
| | laser speed ($mm/s$) | [120, 1530] | [200, 1100] | [300, 600] |
| | hatching space ($\mu m$) | [80, 100] | [40, 120] | [70, 120] |
| | energy ($J$) | [17.99, 150] | [50.62, 292] | [41.7, 98.8] |
| Mechanical property | | Relative density (%) | | |

## 3 Reproduced Modeling Methods

During the case study, five TL methods are selected and integrated with DTR or ANN to form six TL-based AM modeling methods, whose structures and details are presented as follows.

### 3.1 Instance-based TL

As a classical TL-based regression framework, the Two-stage TrAdaBoost.R2 is selected to construct an instance-based decision tree regression (I-DTR) model, which is implemented with the *DecisionTreeRegressor* function in the *Sklearn* library and the open-source code of the Two-stage TrAdaBoost.R2 [48]. The overall procedure is shown below, and more details are discussed in the reference [49].

*Step 1*: Given the source dataset $\boldsymbol{D}^s = \{(\boldsymbol{x}_i^s, y_i^s) | i \in [1, n^s]\}$ and target dataset $\boldsymbol{D}^t = \{(\boldsymbol{x}_i^t, y_i^t) | i \in [1, n^t]\}$, a new dataset is defined as the combination $\boldsymbol{D} = \{\boldsymbol{D}^s, \boldsymbol{D}^t\}$, where the initial weight for each data is identical, i.e., $\boldsymbol{w}^1 = \{w_i^1 = \frac{1}{n^s + n^t}, i \in [1, n^s + n^t]\}$. $n^s$ and $n^t$ are the source data size and the target data size respectively. The global iteration number is set as $J = 1$.

*Step 2*: To reduce the risk of overfitting, a boosting procedure is performed to update the weights of target data. The boosting iteration is set as $j = 1$, and the maximum iteration is $N$.

  *Step 2*-1: At each boosting iteration $j$, a DTR model $f_{DTR}^j$ is constructed based on $\{\boldsymbol{D}, \boldsymbol{w}^J\}$ and validated on the whole dataset $\boldsymbol{D}$. Then a weighted mean error of $f_{DTR}^j$ is defined as $\varepsilon_j = \sum_{i=1}^{n^s+n^t} w_i^J e_i$, where $e_i$ is an adjusted prediction error of data $\boldsymbol{x}_i$. If $\varepsilon_j > 0.5$, the procedure jumps to *Step2*-2; Otherwise, the weight of each target data is updated as $w_i^J = w_i^J \beta^{1-e_i}/Z_J, i \in [n^s + 1, n^s + n^t]$, where $\beta = \varepsilon_j/(1 - \varepsilon_j)$ and $Z_J$ is a normalizing constant. As a result, the target data with large prediction errors are emphasized during modeling in the next iteration $j = j + 1$.

  *Step 2*-2: Given any data $\boldsymbol{x}$, its prediction is the weighted medium of predictions obtained by all DTR models. And the *K*-fold cross-validation framework is used to calculate the prediction error $E_J$ of the final prediction model.

*Step 3*: Based on the same process in *Step2*-1, both the adjusted error $e_i$ and the coefficient $\beta$ are calculated. The weights of source data are updated as $w_i^J = w_i^J \beta^{1-e_i}/Z_J, i \in [1, n^s]$, while the weights of target data are updated as $w_i^J = w_i^J/Z_J, i \in [n^s + 1, n^s + n^t]$. Generally, the weights of relevant source data are increased. Therefore, the possibility of selecting relevant source data would be increased gradually to improve the accuracy of the target model in the next global iteration.

*Step 4*: The procedure returns to *Step 2*, and the global iteration number is updated as $J = J + 1$. When $J$ is large than a defined threshold $maxJ$, the modeling process terminates, and the model with minimum error $E_J$ is defined as the output model.

In this paper, the maximum number of boosting iterations, $N$, the number of maximum global iterations, $maxJ$, and the fold number $K$ in cross-validation in I-DTR are set in Table 3, and all other hyperparameters are set as default values as in [48].

### 3.2 Feature-based TL

Considering different data sizes in the source and target tasks, the simplified transformation matrix integrated subspace alignment method [11] is selected and reproduced. In this method, the simplified transformation matrix [50] is used to transfer the source feature space to the target feature space, shown as

$$H^{s \to t} = (A^T A)^{-1} A^T B \tag{1}$$

where $A = [X^t, f^s(X^t), \mathbf{1}]_{n^t \times (n_d+2)}$, $B = [Y^t]_{n^t \times 1}$, $f^s$ is the pre-trained source model, and $n_d$ is the dimension of the input variable. Then the initial source feature representation $D^{s,0} = [X^s, Y^s, \mathbf{1}]_{n^s \times (n_d+2)}$ is used to construct the transformed source data $D^{s,1} = [X^s, D^{s,0} H^{s \to t}]_{n^s \times (n_d+1)}$, where $D^{s,0} H^{s \to t}$ is regarded as the transformed source outputs. Based on the transformed source data, two modeling frameworks are designed.

- *SA-DTR*: The combination of target data $[X^t, Y^t]$ and transformed source data $D^{s,1}$ is used to construct a DTR model directly, which is denoted as a subspace alignment based DTR model.
- *SA-I-DTR*: Instead of feeding $[X^t, Y^t]$ and $D^{s,1}$ to the conventional DTR modeling process directly, the new method selects I-DTR as the modeling process. In other words, the source data applied in I-DTR, as described in Section 3.1, is replaced by the transformed one $D^{s,1}$. This method is represented as subspace alignment based I-DTR model, which is an application of the combination of instance-based and feature-based TL methods.

From the perspective of training samples, SA-DTR and SA-I-DTR are similar because both target and transformed source data are included. The only difference is that the former is constructed based on all data, while the latter is based on a subset sampled from all data according to their corresponding weights.

### 3.3 Model-based TL

For the model-based TL, two frameworks, e.g., source model transformation and fine-tuning, are selected to form two comparative methods.

- *Source model transformation*

A source DTR model is first constructed based on source data $D^s$. Then the transformation framework in the reference [11] is used to construct the target model by solving the following optimization problem by Genetic Algorithm (GA) in the Python library *geneticalgorithm2* [51],

$$min \; \|M_1 f^s(M_2 X^t + B_2 | \boldsymbol{\theta}^s) + B_1 - Y^t\|_2^2 \tag{2}$$

where $M_1$, $B_1$, and $B_2$ are scalar, and $M_2$ is a vector with size $4 \times 1$ in this paper. Therefore, the optimization problem has in total seven variables. $M_2 X^t$ is the element-wise product between $M_2$ and each target input variable $x^t$. After finding the optimal transformation vectors, the target model $f^t = M_1 f^s(M_2 X^t + B_2) + B_1$ is denoted as the reformulated DTR (Re-DTR) in the following discussion. In this paper, some hyperparameters about GA and the optimization problem are summarized below, and others are set as default values.

- *Fine-tuning framework*

ANN is selected to test the performance of the fine-tuning framework. Figure 1(a) shows the applied ANN structure, where three hidden layers with sizes $8 \times 1$, $16 \times 1$, and $8 \times 1$ are used with a ReLU activation function respectively. The procedure for training a target ANN model is described below.

*Step 1*: The source ANN model is initialized randomly, and trained on source data by minimizing the mean square error (MSE) loss with the Adam optimizer in Pytorch [52].

*Step 2*: The first three layers of the pre-trained source model are shared and frozen in the target ANN model, whose last two layers are initialized as the counterparts in the pre-trained source model and fine-tuned based on limited target data.

The obtained target model is denoted as fine-tuning based ANN (FT-ANN) in this case study. Some hyperparameters applied to train ANN models are summarized in Table 3, including the learning rates $lr^s$ and $lr^t$ for source and target model respectively, the iteration size $ite^s$ for training the source model, and the iteration size $ite^t$ for fine-tuning the target model. Other hyperparameters about ANN structures and Adam optimizer are set as default values in Pytorch [52].

Table 2 Hyperparameters for GA

| Parameter Name | Value |
|---|---|
| Max number of iterations | 5000 |
| Population size | 200 |
| Low bound of design space for each variable | -10 |
| Upper bound of design space for each variable | 10 |

### 3.4 Multi-task learning

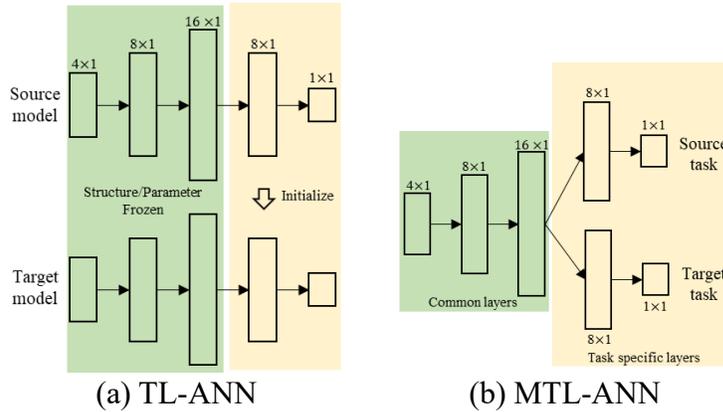

(a) TL-ANN  (b) MTL-ANN
Figure 1 Structures of TL-ANN and MTL-ANN

For the multi-task learning based ANN (MTL-ANN) model, its structure is designed based on the hard parameter sharing mechanism [53] shown in Figure 1(b). The source task and the target task share the input layer and the first two hidden layers, but have two task-specific layers to predict their outputs respectively. Each layer in MTL-ANN shares the same randomly initialized parameters as its corresponding layer in the initialized source model in Section 3.3. In this case study, the loss function $L$ is defined as the sum of MSE losses in source and target tasks without regularization, i.e., $L = l^s + l^t$. The Adam optimizer is also applied to updating MTL-ANN with the learning rate $lr^{MTL}$ and the number of training iterations $ite^{MTL}$, shown in Table 3.

## 4 Validation Framework

As mentioned in Section 2, the main purpose of this paper is to compare the performance improvements caused by different TL methods in AM modeling. For this purpose, a general validation framework is designed in Figure 2, whose details are discussed below.

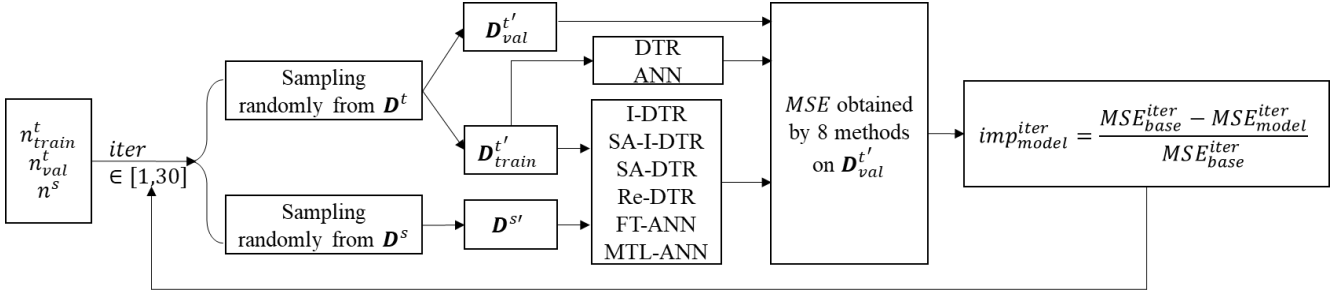

Figure 2 A common validation framework for all models

Based on the raw dataset $\boldsymbol{D}^t$ and $\boldsymbol{D}^s$ in Section 2, the target training data $\boldsymbol{D}_{train}^{t'}$ with size $n_{train}^t$, the target validation data $\boldsymbol{D}_{val}^{t'}$ with size $n_{val}^t$, and the source data $\boldsymbol{D}^{s'}$ with size $n^s$ are sampled randomly at each validation iteration $iter \in [1,30]$. Then six reproduced TL-based models are constructed based on $[\boldsymbol{D}_{train}^{t'}, \boldsymbol{D}^{s'}]$, and the baseline DTR and ANN models are constructed only with $\boldsymbol{D}_{train}^{t'}$ for comparison. The eight constructed models are then validated on $\boldsymbol{D}_{val}^{t'}$ to quantify the improvements in modeling performance caused by each TL framework:

$$imp_{model}^{iter} = \begin{cases} \frac{MSE_{DTR}^{iter} - MSE_{model}^{iter}}{MSE_{DTR}^{iter}} & model \in \{\text{I-DTR, SA-I-DTR, SA-DTR, Re-DTR}\} \\ \frac{MSE_{ANN}^{iter} - MSE_{model}^{iter}}{MSE_{ANN}^{iter}} & model \in \{\text{FT-ANN, MTL-ANN}\} \end{cases} \quad (3)$$

where $MSE_{DTR}^{iter}$ and $MSE_{ANN}^{iter}$ are MSE values of baseline DTR and ANN models at the current iteration respectively, and $MSE_{model}^{iter}$ represents the MSE value of TL-based models. If $imp_{model}^{iter} > 0$, a positive transfer is observed as the MSE value is reduced after applying the TL framework; otherwise, a negative transfer occurs.

Considering the randomness in data sampling and source/target model training processes, the median $imp_{model}^{iter}$ value among 30 iterations, and the positive transfer ratio ($\frac{number\ of\ positive\ transfers}{30}$) are used to reflect the overall performance of each TL framework at the given data size $\{n_{train}^t, n_{val}^t, n^s\}$. Although a large uncertainty would be observed in some cases, this uncertainty could be reduced by optimizing the selection of training data, modeling processes, and hyperparameters, which is not the purpose of this paper. Besides, in the case study, each model is trained only 10 times in the early stage and then another 20 times are added for each model. When comparing the median MSE improvement and positive transfer ratio, the overall trends observed in 10-times training are similar to the ones in 30-times training. In other words, the uncertainty does not affect the analysis of the results too much to some extent. Therefore, the uncertainty will not be discussed in the next section.

## 5  Results and Discussion

In this case study, different combinations of $\{n_{train}^t, n_{val}^t, n^s\}$ are used to study the effects of training data size ratio $n_{train}^t/n^t$ on various TL methods, and two source datasets are applied to compare the effects of similarities. According to the data collected in Section 2, $n_{train}^t$, $n_{val}^t$, and $n^s$ are designed as Eq. (4). Therefore, there are total 27 (9 × 3) combinations of data sizes for target modeling with source task 1, and 15 (5 × 3) combinations for target modeling with source task 2. Given any combination, the designed validation framework is performed to obtain corresponding performance reflectors of each TL framework. Also, two hyperparameter settings shown in Table 3 are selected for training to reduce their effects on comparisons.

$$\begin{cases} n^{s1} \in \{17,21,25,29,33,37,41,45,49\} & for\ \boldsymbol{D}^{s1} \\ n^{s2} \in \{17,21,25,29,32\} & for\ \boldsymbol{D}^{s2} \\ n_{train}^t \in \{8,12,16\}, n_{val}^t = 8 & for\ \boldsymbol{D}^t \end{cases} \quad (4)$$

Table 3 Two hypermeter settings for comparison in reproduced models

| Hyperparameter | Notation | Setting 1 | Setting 2 |
|---|---|---|---|
| Maximum number of booting iterations | $N$ | 100 | 50 |
| Number of maximum global iteration | $maxJ$ | 10 | 10 |
| Fold number in cross-validation | $K$ | 5 | 3 |
| Learning rate for the source model | $lr^s$ | 0.1 | 0.05 |
| Learning rate for the target model | $lr^t$ | 0.1 | 0.05 |
| Learning rate in MTL | $lr^{MTL}$ | 0.1 | 0.05 |
| Iteration size to train the source model | $ite^s$ | 10000 | 10000 |
| Iteration size to fine-tune the target model | $ite^t$ | 5000 | 5000 |
| Iteration size for MTL | $ite^{MTL}$ | 5000 | 5000 |

## 5.1 Similarity

Considering different data sizes in two source tasks, a truncated $n^{s1'} \in \{17, 21, 25, 29, 33\}$ is used to compare TL performances caused by two sets of sources that represent different similarities. Therefore, each model has 15 data points for comparison when trained with each hyperparameter setting and source task.

As mentioned above, the qualitative similarity between source task 2 and the target task is higher in this case study. From the results in Figure 3 (a) and Figure 4 (a), more positive median MSE improvements and smaller absolute values of negative improvements are observed in SA-DTR, Re-DTR, FT-ANN, and MTL-ANN models with source task 2 than those with source task 1 during the entire $n_{train}^t/n^s$ range. Although the median MSE improvements of SA-I-DTR with source 1 is better than source 2 when $n_{train}^t/n^s > 0.6$, both I-DTR and SA-I-DTR trained with source task 1 are comparable with the ones trained with source task 2 in most cases.

Moreover, compared with models with source task 1 in Figure 3 (b), a clear increase in the positive transfer ratio is monitored in I-DTR, FT-ANN, and MTL-ANN with source task 2 at most $n_{train}^t/n^s$ values. And a same phenomenon occurs in Re-DTR, FT-ANN, MTL-ANN with source task 2 and hyperparameter setting 2, shown in Figure 4 (b). All other models shown in Figure 3 (b) and Figure 4 (b) have comparable performances in the positive transfer ratio between two source tasks.

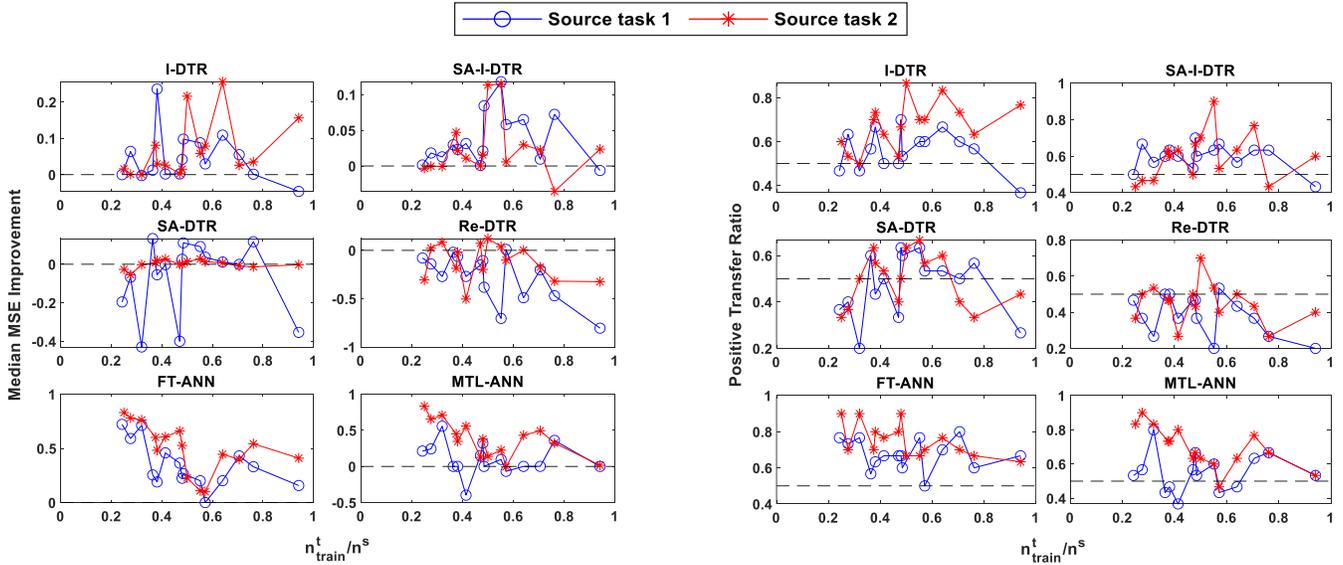

(a) Median $imp_{model}^{iter}$      (b) Positive transfer ratio

Figure 3 Comparison of similarities for TL-based models with hyperparameter setting 1

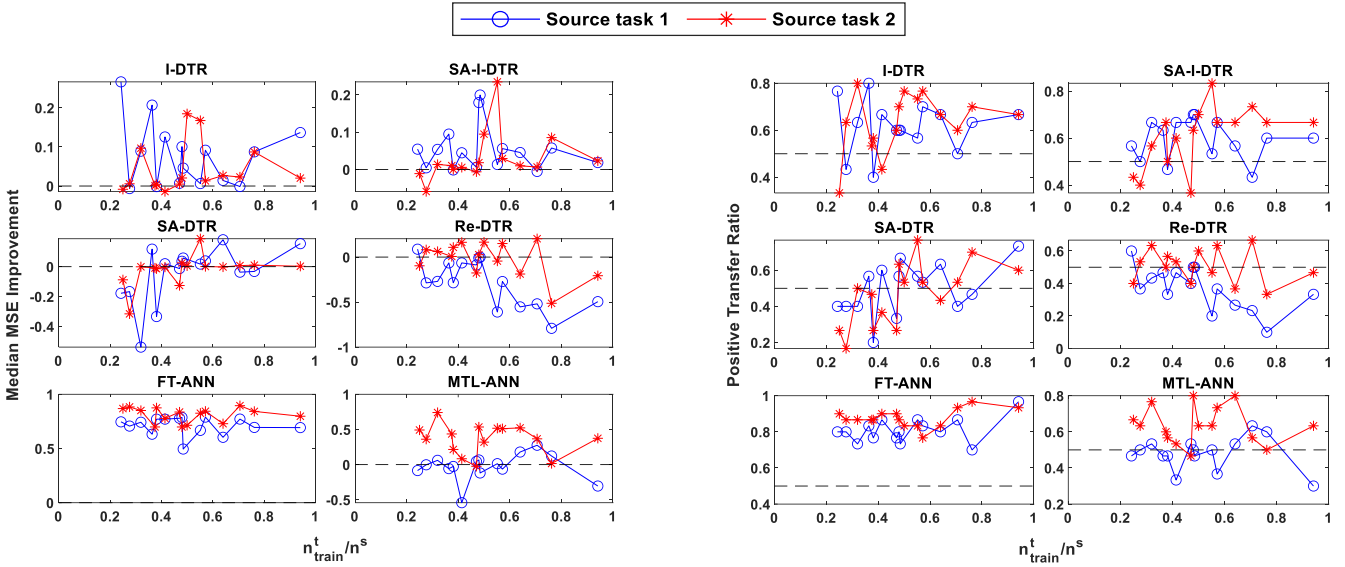

(a) Median $imp_{model}^{iter}$  (b) Positive transfer ratio

Figure 4 Comparison of similarities for TL-based models with hyperparameter setting 2

Comparison results illustrate that the TL method would provide a larger improvement in modeling performance when the source and target domains have higher similarities. Before data collection, quantifying similarities between source and target domains is difficult, since little knowledge about domains is available. Therefore, to find acceptable source data, qualitative similarity could be an alternative. In other words, given a certain target AM problem, collecting source data from products fabricated with geometries, materials, AM processes, or machines within the same series as the target product would be better.

## 5.2 Training data size

In this section, only source task 1 is selected to study the effect of the training data size ratio $n_{train}^t/n^s$, as it provides 27 data points and a wider ratio range $[\frac{8}{49}, \frac{16}{17}]$. The median $imp_{model}^{iter}$ value and the positive transfer ratio at different $n_{train}^t/n^s$ are summarized in Figure 5 and Figure 6. From the perspective of median MSE improvements and positive transfer ratio, three common trends are observed.

First, a positive transfer ratio is found in all TL-based models at all $n_{train}^t/n^s$ values. This demonstrates that all selected TL methods can improve the target modeling performance, which reflects the value of applying TL in AM modeling problems.

When $n_{train}^t/n^s$ is smaller than a threshold (e.g., 0.2 in this case), the median MSE improvements in I-DTR, SA-DTR, SA-I-DTR, and Re-DTR are mostly negative, as shown in Figure 5 (a) and Figure 6 (a). Meanwhile, these models also suffer from negative transfer in most cases, as the positive transfer ratio is smaller than 0.5 in Figure 5 (b) and Figure 6 (b). The reason is that the fewer target training data, the less target information they can provide, resulting in the difficulty of extracting relevant and appropriate knowledge from source data. If inappropriate source knowledge is obtained, it would mislead the construction process and cause negative transfer.

In contrast, when the sizes of target training data and source data are close (e.g., $n_{train}^t/n^s > 0.7$ in this case), a decreasing trend is monitored in both performance reflectors for all models in Figure 5 (a) and Figure 6 (a). This reduction is attributed to the fact that a target model with acceptable accuracy could be constructed based on target data only. Then, the difference between source and target domains would be amplified when transferring source knowledge, posing negative effects on training the target model. Although a performance increase is observed for most models in Figure 5 (b) and Figure 6 (b), generating target training data with a size close to source data indicates many expensive experiments or simulations, which is not preferable in TL-based AM applications.

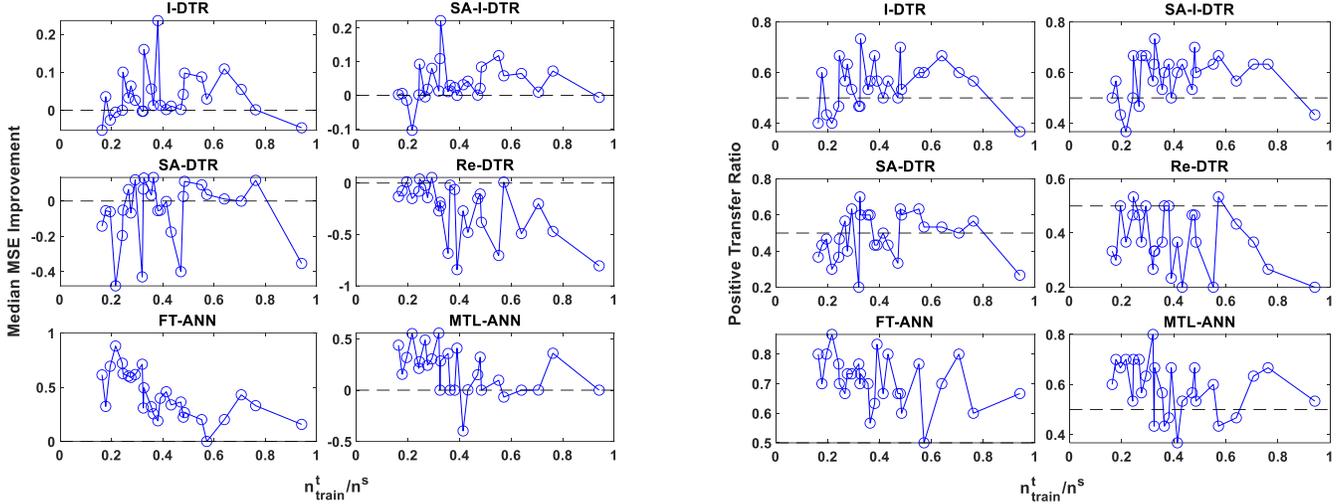

(a) Median $imp_{model}^{iter}$

(b) Positive transfer ratio

Figure 5 Comparison of TL-based models for $n_{train}^t/n^s$ with hyperparameter setting 1

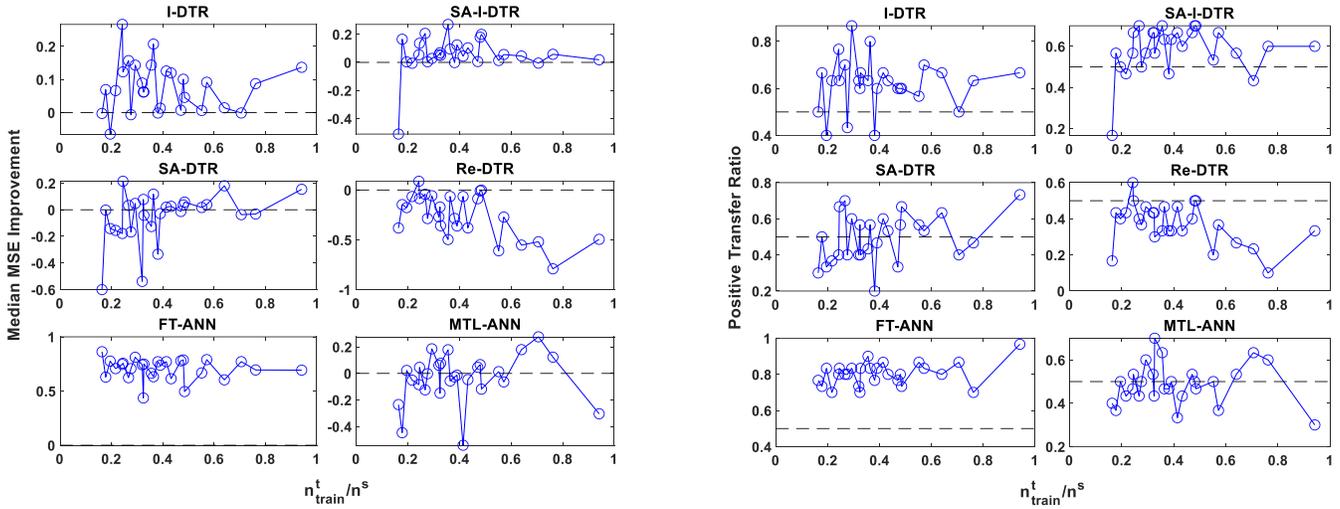

(a) Median $imp_{model}^{iter}$

(b) Positive transfer ratio

Figure 6 Comparison of TL-based models for $n_{train}^t/n^s$ with hyperparameter setting 2

Therefore, in real-world AM problems, the size ratio $n_{train}^t/n^s$ is recommended to be in a certain range (e.g., [0.2,0.7]) to reach a tradeoff among computational expense, modeling accuracy, and TL performances.

### 5.3 Data preprocessing

In the above two comparisons, all models are constructed by raw datasets without data preprocessing. To find the effects of different data preprocessing techniques on TL methods, the min-max scaler ($x' = \frac{x-\min(x)}{\max(x)-\min(x)}$) and Z-score standardization ($x' = \frac{x-\mu}{\sigma}$, where $\mu$ and $\sigma$ are the mean and stand deviation of $x$) [54] are applied to input variables respectively. Considering the preprocessed data would affect the performances of both baseline and TL-based models, their median MSE values are compared in Figure 7 (a)-(b) and Figure 8 (a)-(b).

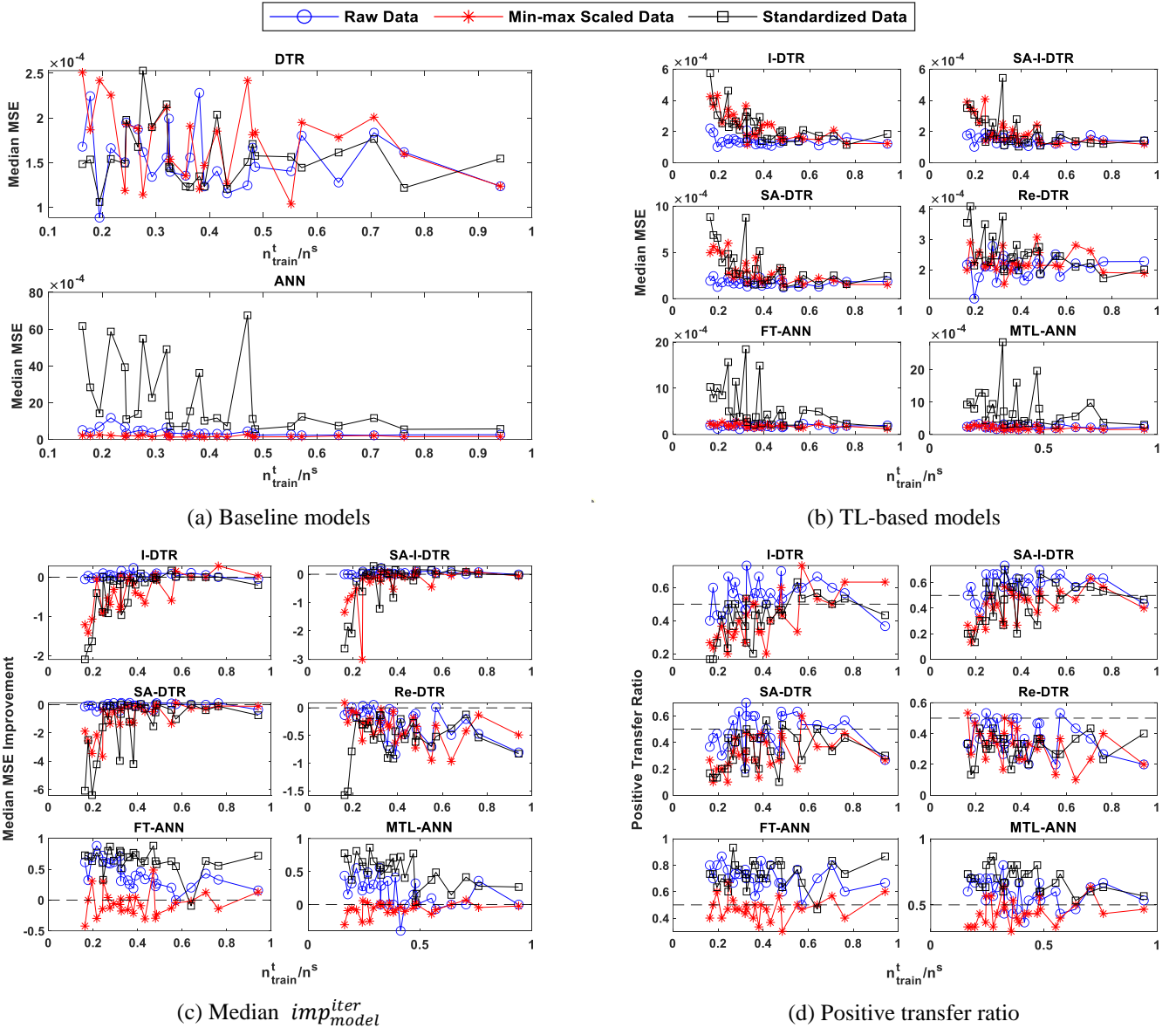

(a) Baseline models  (b) TL-based models

(c) Median $imp_{model}^{iter}$  (d) Positive transfer ratio

Figure 7 Comparison of TL-based models with hyperparameter setting 1 for data preprocessing

Comparing the median MSE in Figure 7 (a) and Figure 8 (a), the baseline DTR models with raw, min-max scaled, and standardized data are comparable, as they capture the smallest median MSE at different $n_{train}^t/n^s$ values respectively. However, I-DTR, SA-I-DTR, SA-DTR, and Re-DTR models all suffer from the applied data preprocessing techniques. For instance, in Figure 7 (b) and Figure 8 (b), the four TL-based DTR models trained with min-max scaled and standardized data have a larger median MSE than those with raw data at most $n_{train}^t/n^s$ values, which indicates a lower accuracy. Also, at some $n_{train}^t/n^s$ values, the data preprocessing techniques may bring about negative transfer problem, as the TL-based DTR models have a larger median MSE than the baseline DTR model. Similarly, better median MSE improvements and larger positive transfer ratios are also observed in the TL-based DTR models with raw data. Especially, the TL-based DTR models with min-max scaled and standardized data are afflicted with the negative transfer problem, as many negative median MSE improvements and positive transfer ratios smaller than 0.5 are found in Figure 7 (c)-(d) and Figure 8 (c)-(d). The main reason for the above problems is that given one source data $(\boldsymbol{x}^s, y^s)$, one target data $(\boldsymbol{x}^t, y^t)$, and $\boldsymbol{x}^s \neq \boldsymbol{x}^t$, the source and target input variables may be close or even the same after preprocessing according to their datasets respectively. Such a pair of preprocessed data would deteriorate modeling performances when transferring source data as knowledge directly (e.g., the combination of source and target data in the instance-based framework), or comparing source and target data directly (e.g., space alignment in the feature-based framework, and space scale/shift in the model-based framework).

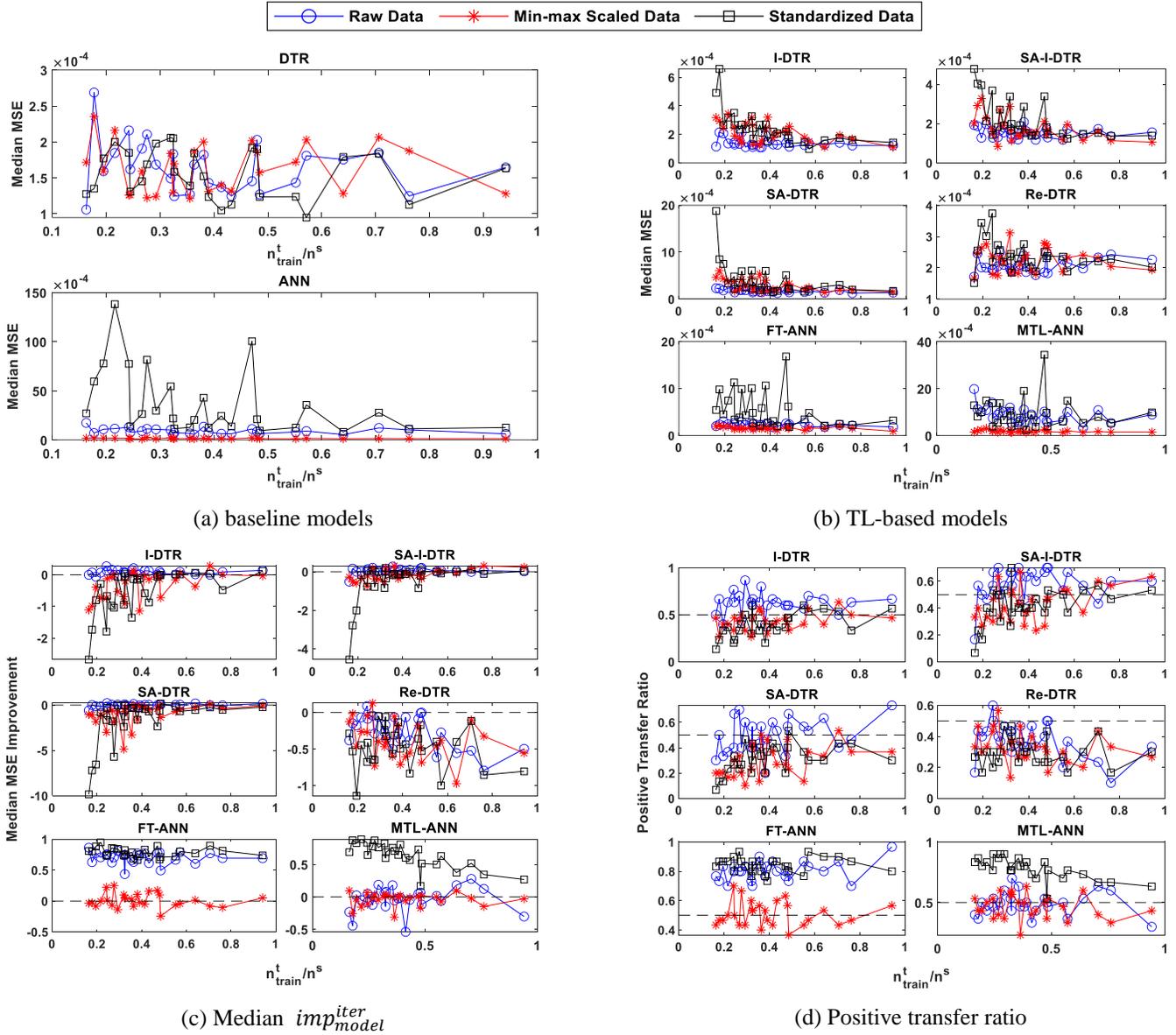

(a) baseline models

(b) TL-based models

(c) Median $imp_{model}^{iter}$

(d) Positive transfer ratio

Figure 8 Comparison of TL-based models with hyperparameter setting 2 for data preprocessing

On the contrary, a different trend is observed in the baseline ANN, FT-ANN, and MTL-ANN models. On the one hand, the min-max scaler can bring better modeling performances for FT-ANN and MTL-ANN in most cases, as shown in Figure 7 (b) and Figure 8 (b). But they also result in the worst improvements caused by TL considering the smallest median MSE improvements and positive transfer ratios in Figure 7 (c)-(d) and Figure 8 (c)-(d). The reason is that the baseline ANN with min-max scaled data has the smallest median MSE. It would be harder to further improve the modeling performance by TL, and a minor reduction of MSE would only provide a small median MSE improvement. On the other hand, although FT-ANN and MTL-ANN models trained with standardized data perform worse than those with raw data, they have the largest median MSE improvement (Figure 7 (c) and Figure 8 (c)) and positive transfer ratios (Figure 7 (d) and Figure 8 (d)) during the entire $n_{train}^t/n^s$ range. This trend is attributed to the large reduction of MSE caused by TL, considering that the baseline ANN trained with standardized data has the largest median MSE shown in Figure 7 (a) and Figure 8 (a).

To sum up, the data preprocessing techniques could be applied in fine-tuning and MTL, as both frameworks transfer implicit source knowledge represented as model parameters and structures. But they should be selected carefully to balance the modeling performance and the performance improvement.

## 6  Summary

This paper designs a case study to shed light on challenges when applying transfer learning in AM modeling, including *which source domain to use, how much target data is needed,* and *whether to apply data preprocessing techniques*. Based on an open-source dataset about laser powder bed fusion and material Ti-6Al-4V, two source tasks and one target task are defined according to different printers in this case study. Then, six transfer learning-based models are reproduced by integrating the decision tree regression and the artificial neural network with five different transfer learning methods, including Two-stage TrAdaBoost.R2, space alignment, source model transformation, fine-tuning, and multi-task learning. To compare these methods, a general validation framework is designed to calculate the MSE improvement caused by transfer learning and the positive transfer ratio among 30 runs. These comparison results are then discussed to explore the effects of similarity, training data size, and data preprocessing on transfer learning. Based on the discussion, three recommendations are provided for applying transfer learning to real-world AM problems.

- Considering the difficulty of calculating the similarity among different AM products, it is better to find source domain data from products with a large qualitative similarity.
- Given certain data about source products, the target training data size should be within a certain range to balance the computational expenses of target data collection, modeling accuracy, and transfer learning performance. The range of target-to-source training data size ratio is recommended as [0.2,0.7].
- For data processing, it is not suggested when the transfer learning frameworks reuse the source data as knowledge or compare source/target data directly. For fine-tuning and multi-task learning framework, the data preprocessing should be determined according to the data and problem characteristics, to reach a trade-off between modeling performance and performance improvement.

**Acknowledgements**

Funding from Natural Science and Engineering Research Council (NSERC) of Canada under the project RGPIN-2019-06601, as well Graduate Dean's Entrance Scholarship (GDES) from Simon Fraser University, is appreciated.